\documentclass[conference]{IEEEtran}
\IEEEoverridecommandlockouts
\usepackage{cite}
\usepackage{amsmath,amssymb,amsfonts}
\usepackage{algorithmic}
\usepackage{graphicx}
\usepackage{textcomp}
\usepackage{xcolor}
\def\BibTeX{{\rm B\kern-.05em{\sc i\kern-.025em b}\kern-.08em
    T\kern-.1667em\lower.7ex\hbox{E}\kern-.125emX}}
    
\usepackage{hyperref}
\usepackage{wrapfig}
\usepackage{subcaption}
\usepackage{hhline}
\usepackage{multirow}
\usepackage{tikz}
\usepackage{arydshln}
\usepackage{booktabs}

\begin{document}

\title{Camera–Radar–LiDAR Fusion via Point-Cloud Projection for Multi-Task Learning}
% \title{Camera–Radar–LiDAR Fusion via Point-Cloud Projection for Multi-Task Learning}
\author{\IEEEauthorblockN{1\textsuperscript{st} Kavin Chandrasekaran}
\IEEEauthorblockA{\textit{Research and Development Department} \\
\textit{Elektrobit Automotive GmbH}\\
Munich, Germany \\
Kavin.Chandrasekaran@elektrobit.com}
\and
\IEEEauthorblockN{2\textsuperscript{nd} Sorin Grigorescu}
\IEEEauthorblockA{\textit{Department of Robotics} \\
\textit{Transilvania University of Brasov}\\
Brasov, Romania \\
s.grigorescu@unitbv.ro}
}

\maketitle

\begin{abstract}
Autonomous vehicles require robust perception systems to operate safely under diverse environmental conditions. While cameras provide high-resolution visual data, their performance significantly degrades in low-light and adverse weather. High Definition (HD) radar offers superior reliability and angular resolution under such conditions but lacks rich visual feature recognition. LiDAR, though expensive, delivers highly precise distance measurements and dense point clouds. Each sensor exhibits unique trade-offs, necessitating the development of advanced sensor fusion algorithms to harness complementary strengths while minimizing computational complexity and maintaining accuracy. In this study, the proposed fusion architecture performs early fusion by combining camera images with perspective-view point cloud data from HD radar and LiDAR sensors. The proposed model is trained to detect vehicles and segment free drivable space simultaneously. Experiments are performed on the RADIal dataset where we demonstrate that fusing point cloud data with camera images in front-view surpasses the performances of the most recent standalone and fusion models.
\end{abstract}


\section{INTRODUCTION}
\label{introthirdpaper}
Establishing a reliable perception is considered very important for any Autonomous Driving (AD) system which relies on sensors like camera, radar, and LiDAR. To ensure reliability and mitigate perception errors, the system requires a suitable design choice which remains an ongoing and challenging research question~\cite{janai_computer_2021}.

Several fusion strategies over the last years have employed camera-LiDAR sensor data for 3D object detection~\cite{cai_bevfusion4d_2023, xia_insfusion_2025} and semantic segmentation~\cite{zhao_lif-seg_2021}. The authors of~\cite{drews_deepfusion_2022, man_bev-guided_2023} use a camera, LiDAR, and radar to perform object detection and segmentation, respectively. The main advantage of using LiDAR with camera data is that it provides denser point clouds, high angular resolution, and range detection accuracy. Camera images, on the other hand, offer rich semantic information of the scene. Despite these advantages, both the camera and the LiDAR face substantial performance degradation in challenging weather conditions and require additional processing power, which are crucial requirements for a stable autonomous driving system over the long term. 

\begin{figure}[!ht] % figure 8166
    \centering % <-- added
    \label{figarch:introoverview_new}
\begin{tikzpicture}
        % Base image
        \node[anchor=north west] (base) at (0, 0) 
            {\includegraphics[width=8.5cm, height=4.2cm]  {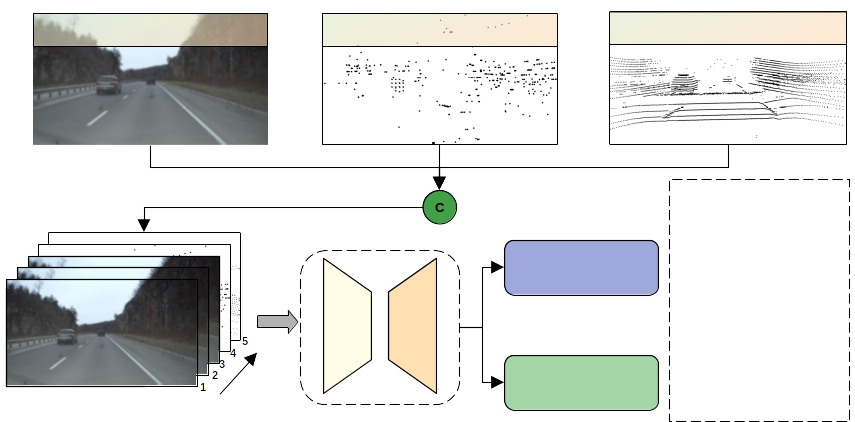}};
  
        \node[anchor=north west] (base) at (6.77, -2) 
            {\includegraphics[width=1.6cm, height=1cm]  {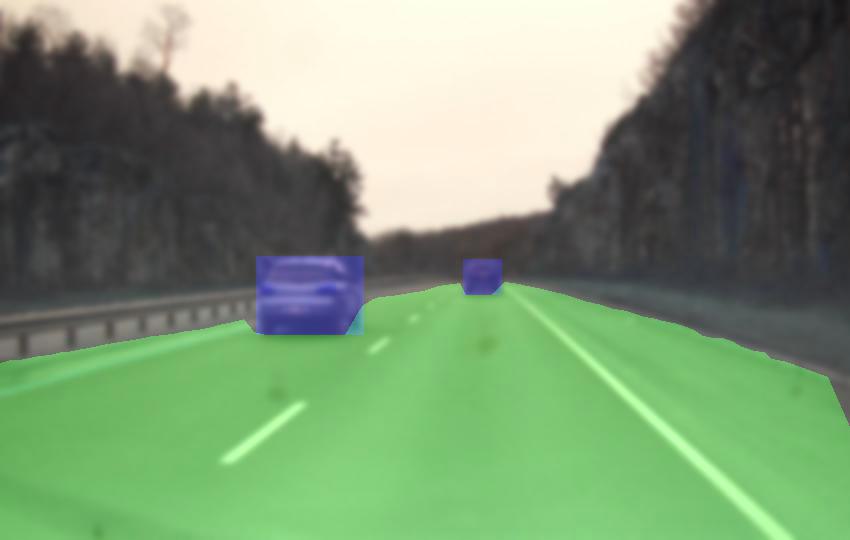}};

        \node[anchor=north west] (base) at (6.77, -3.15) 
            {\includegraphics[width=1.6cm, height=1cm]  {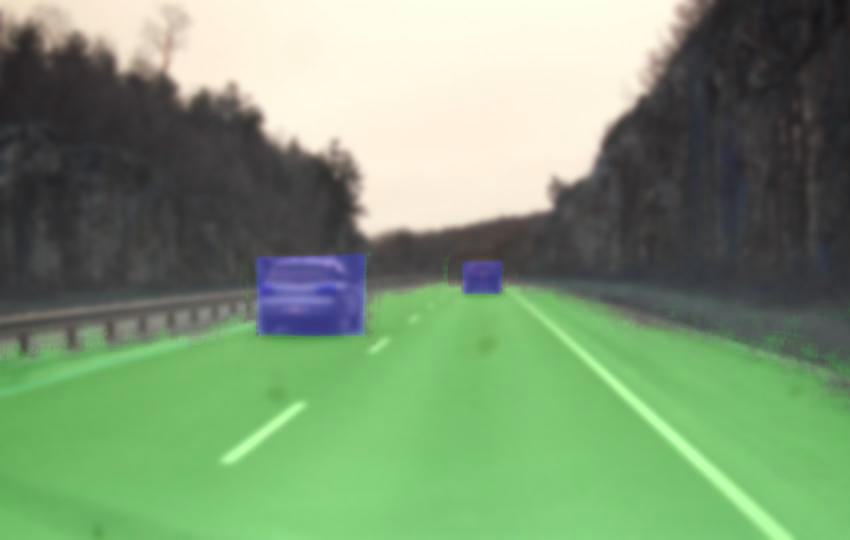}};

        \node[anchor=north] at (1.65, -0.2) {\small{Camera}};
        \node[anchor=north] at (4.5, -0.2) {\small{Radar}};
        \node[anchor=north] at (7.4, -0.2) {\small{LiDAR}};
        \node[anchor=north, rotate=90] at (3.25, -3.35) {\footnotesize{Encoder}};
        \node[anchor=north, rotate=270] at (4.54, -3.4) {\footnotesize{Decoder}};
        \node[anchor=north] at (5.9, -2.55) {\small{Det. Head}};
        \node[anchor=north] at (5.9, -3.7) {\small{Seg. Head}};
        \node[anchor=north] at (7.73, -1.85) {\tiny{Ground-Truth}};
        \node[anchor=north] at (7.73, -3.04) {\tiny{Prediction}};
\end{tikzpicture}      
% \end{subfigure}
\caption{\textbf{Architecture Overview:} Camera-Radar-LiDAR early fusion by concatenation performing vehicle detection and free space segmentation simultaneously. The predicted free drivable space is in green and the detected vehicles in blue.} 
% Green: free space segmentation, orange: vehicle detection.}
\label{fig:introdia}
\end{figure}

Recently, attention has been directed towards the camera-radar data fusion for both object detection~\cite{liu_rofusion_2023, zheng_rcfusion_2023} and semantic segmentation~\cite{rizzoli_multimodal_2022}. This is because automotive radar sensors are affordable and serve as a key component of advanced driver assistance systems in autonomous vehicles. Nonetheless, autonomous driving demands significantly higher levels of safety and robustness, which conventional radar sensors cannot provide fully due to their limited angular resolution. To overcome this drawback, High Definition (HD) imaging automotive radars have been introduced. These systems, which are less susceptible to adverse weather, employ dense antenna arrays to sense both horizontal and vertical dimensions, thereby offering enhanced vertical resolution. However, using raw radar data for fusion is resource inefficient. Hence, in this work we used the point cloud information obtained from HD radar sensor and the denser LiDAR point cloud data. There is still little work on combining camera and LiDAR with HD radar point clouds for deep-learning-based object detection and semantic segmentation, mainly due to data complexity and the scarcity of public annotated datasets.

% A further set of intrinsic constraints in current fusion approaches arises from mismatched feature representations across heterogeneous sensors~\cite{yeong_exploring_2025}. While cameras deliver 2D RGB imagery, LiDAR outputs 3D point clouds and HD radar provides coarse tensors encoding power information. Consequently, each modality produces distinct feature descriptions for the same physical object, which complicates achieving coherent fusion. A promising direction to mitigate this issue is to transform sensor data into a common representational domain, enabling a more robust fusion.

Compared to our recent architectures~\cite{chandrasekaran_resource_2024, chandrasekaran_refnet_2025}, which employ late fusion in Bird’s-Eye View on the RADIal dataset~\cite{rebut_raw_2022}, this work brings the following new contributions:

% Unlike our previous architecture~\cite{chandrasekaran_resource_2024} that proposed late fusion in Bird's-Eye View, in this work, our contributions are threefold:
% In this work, our contributions are threefold:
\begin{itemize}
\item We propose a fusion architecture (Fig. \ref{fig:introdia}) that fuses camera image, HD radar, and LiDAR point cloud data through early concatenation to simultaneously perform object detection and free space segmentation.
\item We demonstrate that fusing camera imagery with HD radar and LiDAR point clouds yields state-of-the-art results on the RADIal dataset~\cite{rebut_raw_2022}.
\item We investigate the trade-offs between accuracy, resource efficiency and computational cost that are often overlooked in prior works. The code is made publicly available via \href{https://github.com/Kavinchandra/crlfusion}{https://github.com/Kavinchandra/crlfusion}.
\end{itemize}

\section{RELATED WORK} \label{relatedwork}
\subsection{Camera-Radar-LiDAR dataset} \label{relatedwork_dataset}

% High-quality fusion datasets that are large-scale and precisely aligned in both time and space are crucial for training deep learning models. Recently, there has been growing interest in exploiting raw radar measurements~\cite{giroux_t-fftradnet_2023, yang_adcnet_2023, jin_cross-modal_2023, chandrasekaran_resource_2024}, although the additional cost of fusing these data with camera images to improve segmentation~\cite{wu_sparseradnet_2024} and detection performance~\cite{liu_rofusion_2023, liu_echoes_2023} is often disregarded.

% Accordingly, various works have provided different radar data representations, including raw Analog-to-Digital Converter (ADC) signals~\cite{roldan_deep_2024}, Range-Azimuth-Doppler (RAD) cubes~\cite{ouaknine_carrada_2021}, Range-Azimuth (RA) maps~\cite{sheeny_radiate_2021}, range-Doppler (RD) spectra~\cite{rebut_raw_2022}, and point clouds (PC)~\cite{paek_k-radar_2023, roldan_deep_2024}. Comprehensive comparisons of these datasets are available in~\cite{yao_radar-camera_2023, wang_multi-sensor_2020}. Because joint use of LiDAR, HD radar, and camera imposes considerable computational and hardware demands, the majority of prior studies have focused on fusing only camera and radar modalities.

Our objective is to perform both free-space segmentation and object detection in a resource-efficient manner. Consequently, we do not consider raw sensor data in this study, as their processing is computationally expensive. Instead, our network operates on camera images, point clouds derived from HD radars, and LiDAR measurements. For that reason, we focus on datasets that provide these modalities, namely RADIal~\cite{rebut_raw_2022}, K-Radar~\cite{paek_k-radar_2023}, and RaDelft~\cite{roldan_deep_2024}. The K-Radar and RaDelft datasets in particular offer 4D radar data collected under a variety of challenging weather conditions; however, analyzing the impact of adverse weather is beyond the scope of this work.

RADIal~\cite{rebut_raw_2022} is chosen as the primary dataset for this study because it offers a comprehensive suite of radar modalities, ranging from raw ADC measurements to point cloud representations, and is tightly synchronized with camera, LiDAR, and odometry data. This rich multimodal configuration enables extensive exploration of various sensor fusion strategies. While the current work focuses on car detection and free space segmentation, the proposed approach can be generalized to additional object categories, contingent on the availability of suitably annotated datasets.

\subsection{Object detection via Camera-Radar-LiDAR fusion} \label{relatedwork_crmethods_det}

% Existing work primarily examines the fusion between cameras and 4D radar or LiDAR~\cite{xiong_lxl_2023, liu_bevfusion_2023}, as well as between 4D radar and LiDAR~\cite{chae_towards_2024, huang_l4dr_2025}. Fusion strategies span from directly combining front-view camera images with 4D radar or LiDAR measurements without explicit view transformation~\cite{vora_pointpainting_2020}, to approaches that first learn a bird’s-eye view (BEV) representation via differentiable transformations~\cite{philion_lift_2020} and then concatenating at the BEV level~\cite{liang_bevfusion_2022, zheng_rcfusion_2023}. More recently, an availability-aware Sensor Fusion (ASF) framework has been introduced~\cite{paek_availability-aware_2025}, which integrates camera, radar, and LiDAR information from the complex K-Radar dataset~\cite{paek_k-radar_2023}.

Recent studies increasingly combine camera imagery with radar point clouds from conventional sensors~\cite{zheng_rcfusion_2023, xiong_lxl_2023, bansal_radsegnet_2022}. FUTR3D~\cite{chen_futr3d_2023} integrates camera images, radar points, and LiDAR voxels from the nuScenes dataset~\cite{caesar_nuscenes_2020}. Other architectures also support this tri-modal setup~\cite{drews_deepfusion_2022, man_bev-guided_2023, hendy_fishing_2020}, but none have been assessed on the HD RADIal dataset~\cite{rebut_raw_2022}. In contrast, while some approaches exploit only radar~\cite{giroux_t-fftradnet_2023, yang_adcnet_2023} from RADIal, our work explicitly targets the joint use of all three modalities. 

A recent survey~\cite{shi_radar_2024} reviews a variety of camera–radar detection architectures, contrasting early~\cite{lei_hvdetfusion_2023}, late~\cite{long_radiant_2023}, and multi-level fusion~\cite{kim_rcm-fusion_2024} strategies. But the models that exploit point cloud data from HD radar and LiDAR with camera images are our focus. As such methods are largely absent, we focus on closely related recent approaches like ROFusion~\cite{liu_rofusion_2023}, Cross Modal Supervision (CMS)~\cite{jin_cross-modal_2023}, EchoFusion~\cite{liu_echoes_2023}, REFNet~\cite{chandrasekaran_resource_2024}, and REFNet++~\cite{chandrasekaran_refnet_2025}.

ROFusion~\cite{liu_rofusion_2023} utilizes a point-wise fusion mechanism that integrates image and range-Doppler (RD) features based on radar points located within annotated bounding boxes. CMS~\cite{jin_cross-modal_2023} instead operates directly on the RD spectrum and depends on pseudo-labels obtained from camera images. EchoFusion fuses Range-Time (RT) data with camera images in BEV through a polar-aligned attention module~\cite{liu_echoes_2023}. Our previous method, REFNet~\cite{chandrasekaran_resource_2024} adopts a separate image processing branch that transforms camera features into a BEV polar representation, which is subsequently combined with range-angle (RA) features from the radar decoder. Our updated REFNet++~\cite{chandrasekaran_refnet_2025} introduces a comparable architecture based on a variational encoder-decoder~\cite{lu_monocular_2019} framework that learns to transform front-view camera data into BEV. In contrast to these approaches, we abstain from using raw radar data such as ADC, RAD, or RD to reduce associated computational and hardware costs. Rather, we encode radar and LiDAR information as binary masks projected into the image domain.

% Section~\ref{method} presents the proposed fusion network, and Section~\ref{experiments} outlines the evaluation protocol. In contrast to much of the prior work, which provides limited analysis of computational demands, our study systematically evaluates both predictive performance and resource consumption. A comprehensive comparison of accuracy and computational efficiency is presented in Section~\ref{results} with qualitative results in Section~\ref{resultsquali}.

Section~\ref{method} presents the proposed fusion network, and Section~\ref{experiments} outlines the evaluation protocol. Unlike previous works with limited computational analysis, our study systematically evaluates both predictive performance and resource consumption. A comprehensive comparison of accuracy and computational efficiency is presented in Section~\ref{results} with qualitative results in Section~\ref{resultsquali}.

% Section~\ref{method} presents the proposed fusion network, and Section~\ref{experiments} details the evaluation protocol. Unlike prior work with limited computational analysis, we systematically assess both predictive performance and resource consumption. A comprehensive comparison of accuracy and computational efficiency is presented in Section~\ref{results} with qualitative results in Section~\ref{resultsquali}.

% Section~\ref{results} compares accuracy and efficiency, with qualitative results in Section~\ref{resultsquali}.

% In this work, as detailed in Section~\ref{results}, we compare the performance of our approach with other recent models not only in terms of accuracy but also in terms of resource consumption. 

% Considering computational complexity metrics is crucial when using raw radar data. Nevertheless, 

\subsection{Segmentation via Camera-Radar-LiDAR fusion}\label{relatedwork_crmethods_seg}
% A taxonomy of algorithms~\cite{rebut_raw_2022, giroux_t-fftradnet_2023, yang_adcnet_2023, jin_cross-modal_2023, wu_sparseradnet_2024} performs free space segmentation and detection in multitasking mode. Methods such as Occupany grid map learning~\cite{jin_semantic_2024} and TransRadar~\cite{dalbah_transradar_2024} focus mainly on free space segmentation using only the raw range-Doppler (RD) data. Quite surprisingly, there exists no camera-radar-LiDAR fusion method specifically designed for free space segmentation using RADIal dataset due to its complex nature. 

A range of existing approaches~\cite{rebut_raw_2022, giroux_t-fftradnet_2023, yang_adcnet_2023, jin_cross-modal_2023, wu_sparseradnet_2024} addresses free-space segmentation in a multi-task setting, simultaneously performing free-space segmentation and object detection. Other lines of work, including occupancy grid map learning~\cite{jin_semantic_2024} and TransRadar~\cite{dalbah_transradar_2024}, concentrate primarily on free-space segmentation from raw range-Doppler (RD) spectrum alone. As far as we are aware, however, no existing method leverages joint camera–radar–LiDAR fusion tailored specifically to free-space segmentation on the RADIal dataset, likely due to the dataset’s inherent complexity.

\section{METHOD} \label{method}

\subsection{Problem statement} \label{problemstatement}
The problem addressed in this work is the design of a unified camera–radar–LiDAR perception architecture that can handle the proposed perception tasks simultaneously and with high resource efficiency, enabled by a point-cloud projection approach.

\subsection{Point cloud projection} \label{pcprojection}
% We are given a set of $N$ camera input images with training sample index $n$, \{$I_{img}^{n} \in \mathbb{R}^{H \times W \times 3}$\}$_{n=1}^{N}$. Given the camera intrinsic parameters $(\Pi \in \mathbb{R}^{3 \times 3})$, height at which the camera is mounted $(h)$ and pitch value $(p)$, the image formation is, $\varepsilon$. 
% \begin{align}
%     \varepsilon = f(\Pi,h,p)
% \end{align}

We consider a collection of $N$ camera images ($I_{img}$), indexed by $n$, denoted as $\{I_{\text{img}}^{n} \in \mathbb{R}^{H \times W \times 3}\}_{n=1}^{N}$. Given the camera intrinsic matrix ($K \in \mathbb{R}^{3 \times 3}$), the mounting height $h$, and the pitch angle $p$, the resulting image formation process is represented by $\Psi$, defined as
\begin{align}
    \Psi = f(K, h, p)
\end{align}

\textit{\textbf{Radar projection}}: Camera-captured images are usually depicted in a perspective view, whereas the point cloud generated by the HD radar sensor is described as \( q_i^{(R)} = (R_i, A_i, E_i, P_i) \), where \( R_i \), \( A_i \), \( E_i \), and \( P_i \) denote the range, azimuth, elevation, and power, respectively. Consequently, defining a common representation is essential to enable effective sensor data fusion for downstream tasks.

% To first align the radar point cloud data, we rotate~\cite{noauthor_rotation_nodate} the points using the sensors pitch, roll and yaw measurements, R$_{\omega}$. The corrected radar coordinates are as follows:

% \begin{equation}
% q_i'^{(R)} =
% \mathbf{R}_{\omega}
% \begin{bmatrix}
% R_i \\
% A_i \\
% E_i
% \end{bmatrix}
% =
% \begin{bmatrix}
% R_i' \\
% A_i' \\
% E_i'
% \end{bmatrix}
% \end{equation}

% The obtained radar points are further mapped into a Cartesian-like representation and projected~\cite{noauthor_opencv_nodate} to camera pixel space using camera's intrinsic and extrinsic parameters encoded in a function as shown below:

% \begin{equation}
% q_i''^{(R)} =
% f_{points}^{pixel}(q_i'^{(R)})
% \end{equation}

To initially align the radar point cloud, we apply a rotation~\cite{noauthor_rotation_nodate} based on the sensor’s pitch, roll, and yaw measurements, denoted by $\mathbf{R}_{\omega}$. The resulting compensated radar coordinates are given by:
\begin{equation}
q_i'^{(R)} =
\mathbf{R}_{\omega}
\begin{bmatrix}
R_i \\
A_i \\
E_i
\end{bmatrix}
=
\begin{bmatrix}
R_i' \\
A_i' \\
E_i'
\end{bmatrix}
\end{equation}

Subsequently, these corrected radar points are converted into a Cartesian coordinate system and projected~\cite{noauthor_opencv_nodate} onto the camera image plane by means of the camera’s internal and external parameters, encapsulated in the function:
\begin{equation}
R_{projected}=
f_{points}^{pixel}(q_i'^{(R)})
\end{equation}

$R_{projected}$ corresponds to radar points projected on the camera pixel coordinates ($u_i, v_i$). Only projected points satisfying the following criteria are retained:
\begin{equation}
\label{boundarycond}
0 \le u_i < W, \quad 0 \le v_i < H
\end{equation}
where, W is the image width and H is the image height.

\textit{\textbf{LiDAR projection:}} The LiDAR point cloud is represented as \(q_i^{(L)} = (r_i, \theta_i, \phi_i, \ell_i)\), where \(r_i\), \(\theta_i\), \(\phi_i\), and \(\ell_i\) denote the radial distance, elevation angle, azimuth angle, and return intensity, respectively.

To correctly align the LiDAR point cloud information, the installation height of the sensor must be taken into account, since the LiDAR unit is mounted beneath the front ventilation grille of the vehicle~\cite{rebut_raw_2022}. After compensating for the elevation angular offset [\(\theta_i' = f(\theta_i)\)] and for the sensor height \(h\), each LiDAR return is transformed into Cartesian coordinates as:
\begin{equation}
\begin{aligned}
x_i^{(L)} &= r_i \cos(\theta_i') \cos(\phi_i) \\
y_i^{(L)} &= r_i \cos(\theta_i') \sin(\phi_i) \\
z_i^{(L)} &= r_i \sin(\theta_i') + h
\end{aligned}
\end{equation}

These 3D points are subsequently mapped into the camera coordinate system and projected~\cite{noauthor_opencv_nodate} onto the image plane:
\begin{equation}
L_{\text{projected}} =
 f_{\text{points}}^{\text{pixel}}\!
\left(
\begin{bmatrix}
x_i^{(L)} \\
y_i^{(L)} \\
z_i^{(L)}
\end{bmatrix}
\right)
\end{equation}
where, \(L_{\text{projected}}\) denotes the LiDAR returns expressed in pixel coordinates \((u_i, v_i)\). Only those projected points that satisfy the condition in Eq.~\ref{boundarycond} are preserved for training.

\subsection{Architecture design} \label{crlfusionnetarchi}
\begin{figure*}
    \centering % <-- added
    \includegraphics[width=0.98\textwidth]{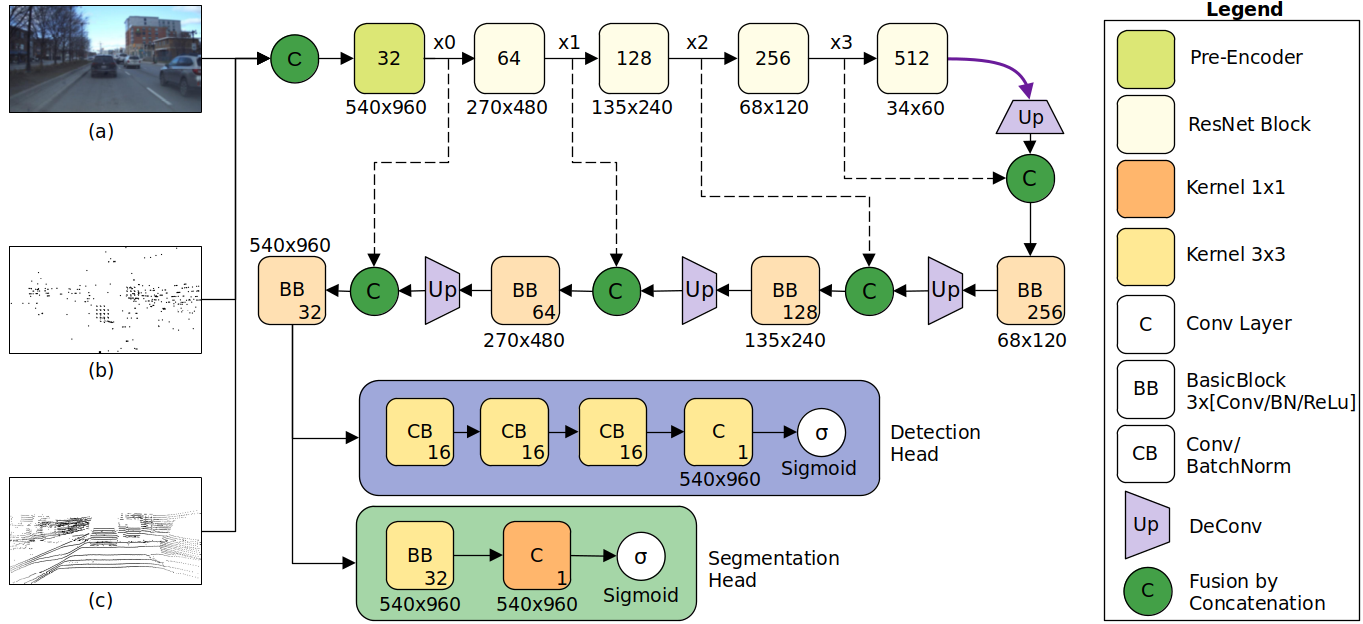}
    % \put(-150.5,55){\rotatebox{90}{\fontsize{10}{10}\selectfont Legend Box}}
    \caption{\textbf{Deep Seeing Through Sensors:} Early fusion is performed using inputs from (a) the camera, (b) a sparse radar point cloud, and (c) a dense LiDAR point cloud. The encoder comprises a pre-encoder followed by four ResNet-50-style blocks, whose output feature maps are denoted as x0, x1, x2, and x3. A thick purple arrow indicates the connection from the encoder to the decoder, through which the encoded features are upsampled to higher spatial resolutions. The dashed lines indicate skip connections that preserve spatial information. The final outputs are produced by a Detection Head and a Segmentation Head, and the corresponding prediction results are illustrated in Fig.~\ref{fig:introdia} and Fig.~\ref{fig:qualitativeresults_fusion_new}.}
    \label{figarch:earlyfusionperspective}
\end{figure*}

Fig.~\ref{figarch:earlyfusionperspective} shows our architectural design, detailing the fusion network with the detection and segmentation head.

\begin{table*}[!h]
\caption{Quantitative results on the test split. ADC, RD, RT, RPC, LPC, C, C\textsuperscript{BEV} correspond to Analog-To-Digital Converter signal, Range-Doppler, Range-Time, Radar Point Cloud, LiDAR Point Cloud, Camera data in Front and Bird's-Eye View respectively. The single modality methods are separated from multimodal fusion methods by a dashed line. Best results are in bold and second best are underlined. The values missing are marked by a "-", either unreported in the respective works or due to code unavailability\dag. \textsuperscript{*}Trained using \textit{camera only} data of the proposed architecture.}
\label{table:quantresul}
\centering
\renewcommand{\arraystretch}{1}
\begin{tabular}{llllllllll}
\hline
% \rule{0pt}{0ex}\\[-1.7ex]
\textbf{} & \textbf{} & \multicolumn{4}{c}{\multirow{1.35}{*}{\centering{Accuracy}}} & \multicolumn{4}{c}{\multirow{1.35}{*}{\centering{Resource Efficiency}}}

\\\cmidrule(lr){3-6} \cmidrule(lr){7-10}

% \textbf{} & \textbf{} & \multicolumn{3}{c}{\multirow{1.35}{*}{\centering{Det.}}} & \multicolumn{1}{c}{\multirow{1.35}{*}{\centering{Seg.}}} & \textbf{} & \textbf{} & \textbf{} 

% \\\cmidrule(lr){3-5} \cmidrule(lr){6-6}

Methods & Modality & AP(\%)$\uparrow$ & AR(\%)$\uparrow$ & F1(\%)$\uparrow$ & mIoU(\%)$\uparrow$ & \# $\downarrow$ & \begin{tabular}[c]{@{}c@{}}Avg \\ FPS\end{tabular} $\uparrow$ & \begin{tabular}[c]{@{}c@{}}Model\\size\end{tabular} $\downarrow$ & \begin{tabular}[c]{@{}c@{}}GPU \\cost\end{tabular} $\downarrow$ \\
\hline
\rule{0pt}{0ex}\\[-1.7ex]
FFTRadNet~\cite{rebut_raw_2022}             & RD                     & 93.45        & 83.35        & 88.11             & 74           & \underline{3.28}              & 67.32       & \textbf{39.2}  & 2.01                                                    \\
TFFTRadNet~\cite{giroux_t-fftradnet_2023}              & ADC                             & 90.80         & 88.31         & 89.54              & 80.2            & 9.13              &52.41         & 109.5    & 2.04                                              \\
ADCNet\dag~\cite{yang_adcnet_2023}            & ADC                                 & 95          & 89           & 91.9              & 78.59            & -             & -          &   -     & -                                                          \\
SparseRad\dag~\cite{wu_sparseradnet_2024}      & RD                & 96.00      & 91.78              &93.84        & 78.48        & -               & -        &  - & -             \\

Occugrid\dag~\cite{jin_semantic_2024}      & RD   & -       & -       & -         & 82.2       & -              & -      & -   & -          \\

TransRadar~\cite{dalbah_transradar_2024}      & RD  & -       & -     & -       & 81.93        & 3.31             & 69.23      & 40.6    & 2.02         \\
 
% \hline
\hdashline
\rule{0pt}{0ex}\\[-1.7ex]
CMS\dag~\cite{jin_cross-modal_2023}            & RD\&C                      & 96.9         & 83.49        & 89.69            & 80.4            & -                 & -        & -   & -                                             \\ 
ROFusion~\cite{liu_rofusion_2023}      & RD\&RPC\&C                      & 91.13        & \textbf{95.29}        &{93.16}        & -    & 3.33            & 56.11             & 87.2            & 2.87                                                        \\ 
EchoFusion~\cite{liu_echoes_2023}      & RT\&C                            & \underline{96.95}        & {93.43}        & \underline{95.16}             & -            & 25.61              & 38.32      & 102.5     & 4.47           \\

REFNet~\cite{chandrasekaran_resource_2024}      & RD\&C\textsuperscript{BEV}       & 95.97        & 87.32        & 91.44           & {84.84}        & 7.28              & 53.05      & 88.2    & 2.06          \\

REFNet++~\cite{chandrasekaran_refnet_2025}      & RD\&C      & 96.16       & 89.43        & 92.67         & \underline{87.58}       & 7.26              & 54.56     & 88.3    & 2.06          \\
\hline

\rule{0pt}{0ex}\\[-1.7ex]
Ours\textsuperscript{*}     & C                & 88.32      & 85.69      & 86.98      & 83.12 & \textbf{3.27}             & \textbf{73.21}      & \underline{39.8}   & \textbf{1.83}     \\
Ours      & C\&RPC\&LPC                              & \textbf{97.32}       &   \underline{94.90}        & \textbf{97.09}             & \textbf{91.77}          & {3.29}              & \underline{70.92}       & 40.4    & \underline{1.97}     \\
\rule{0pt}{0ex}\\[-2.3ex]
\hline
\end{tabular}
\end{table*}

\textit{\textbf{Multi-task learning:}} As detailed in Section~\ref{pcprojection}, radar- and LiDAR-derived point clouds are first projected onto the front-view image plane. The resulting projections are then fused with the RGB camera image (3\texttimes540\texttimes960) by channel-wise concatenation, as illustrated in Fig.~\ref{figarch:earlyfusionperspective}, resulting in a five-channel input representation. The \textit{Pre-Encoder} block initiates the processing pipeline with a standard 3\texttimes3 convolutional layer to perform low-level feature extraction. Subsequently, the Feature Pyramid Network (FPN) encoder comprises four stages that contain 3, 6, 6, and 3 residual layers. Each encoder stage applies 2\texttimes2 downsampling, resulting in an overall reduction in spatial resolution by approximately a factor of 16 in both width and height. At the same time, the number of feature channels is doubled at every downsampling step. This progressive reduction in spatial dimensions is intended to concentrate representational capacity on semantically salient structures.

% in that order, following the design in~\cite{he_deep_2016}
The decoder mirrors this process by progressively recovering spatial resolution via four transposed convolutional layers, each with a 2\texttimes2 kernel and stride of 2, which simultaneously halves the number of feature channels at each stage. To facilitate accurate reconstruction of spatial details, the decoder concatenates features from the deepest encoder layer with multi-scale FPN feature maps (x0--x3) through skip connections, indicated by the dotted arrows in Fig.~\ref{figarch:earlyfusionperspective}. Within each decoding stage, the basic blocks (BB) refine and upsample the features using three consecutive Convolution\texttimes BatchNormalization\texttimes ReLU operations.

The final decoded feature maps are fed into dedicated detection and segmentation heads, which perform object detection and free-space segmentation, respectively. The backbone module is designed to be modular and can be substituted with more computationally intensive architectures when resources permit, enabling a balance between accuracy and efficiency.

\textit{\textbf{Detection Head:}} The detection head processes the shared feature maps using a stack of three Convolution-BatchNormalization blocks, each comprising 16 filters, as illustrated in Fig.~\ref{figarch:earlyfusionperspective}. The final classification stage is implemented by a 3\texttimes3 convolution layer succeeded by a sigmoid activation, which produces a dense probability map indicating, for every pixel, the likelihood of being occupied by a vehicle. Given that the vast majority of pixels correspond to the background, we employ the focal loss~\cite{lin_focal_2018} on the classification branch to address the severe foreground–background imbalance and to stabilize optimization. 

\textit{\textbf{Segmentation Head:}} The segmentation head employs a single basic block (BB) consisting of three consecutive Conv–BatchNorm–ReLU units to refine the decoded feature maps. A subsequent 1\texttimes1 convolution layer generates a 2D prediction map, from which a sigmoid activation estimates the per-pixel probability of drivable versus non-drivable regions. This head is optimized using a binary cross-entropy (BCE) loss function as originally proposed~\cite{rebut_raw_2022}.

\section{PERFORMANCE EVALUATION} \label{experiments}

% \subsection{Dataset} \label{radialdataset}

% RADIal dataset~\cite{rebut_raw_2022} provides well synchronized multimodal data from camera, HD radar, and LiDAR, enriched with GPS information. Following an initial collection of approximately 25,000 temporally aligned frames across the three sensing modalities, a subset comprising 8,252 frames, corresponding to 9,550 annotated vehicles, is labeled. However, the quality of vehicle detection annotations and segmentation masks is constrained by the automated labeling pipeline adopted in~\cite{rebut_raw_2022} and by errors introduced when projecting radar measurements from their native polar coordinate system into the image plane.

\subsection{Training details} \label{traindetails}
The dataset was randomly partitioned, allocating 70\% of the samples for training and approximately 15\% each for validation and testing. We trained the model for 100 epochs using the Adam optimizer~\cite{kingma_adam_2017} with a mini‑batch size of 4. The initial learning rate was set to $1$\texttimes$10^{-4}$ and exponentially decayed by a factor of 0.9 every 10 epochs. The network was trained on a workstation equipped with an NVIDIA RTX A6000 GPU, and 52 GB of RAM.

\subsection{Evaluation metrics} \label{evalmetrics}

\textbf{\textit{Accuracy:}} Average Precision (AP) and Average Recall (AR) evaluated at an Intersection-over-Union (IoU) threshold of 0.5 are adopted as the primary metrics for evaluating object detection performance. The F1 score is subsequently derived from AP and AR using the harmonic mean, defined as \( F1 = \frac{2 \times \text{AP} \times \text{AR}}{\text{AP} + \text{AR}} \). For the free-space segmentation task, performance is quantified using the mean Intersection-over-Union (mIoU), which quantifies how much the predicted regions intersect with the ground-truth regions.
% which measures the overlap between the predicted and ground-truth regions.

% For the free-space segmentation task, performance is measured using the mean Intersection-over-Union (mIoU), which evaluates the overlap between the predicted regions and the ground-truth regions.

\textbf{\textit{Resource efficiency:}} Table~\ref{table:quantresul} also summarizes the resource efficiency of the evaluated methods. The symbol `\#` indicates the total number of learnable parameters, reported in millions. For every frame in the test set, we measure the processing throughput in Frames per Second (FPS) and report the average FPS over all frames. In addition, we provide a comparison of the on-disk model size (in MB) and peak GPU memory usage (in GB), following the evaluation protocol from~\cite{chandrasekaran_resource_2024, chandrasekaran_refnet_2025}.

\subsection{Baselines}
\label{baselines}

The majority of the approaches discussed in Sections~\ref{introthirdpaper} and~\ref{relatedwork} rely on conventional low-resolution radar sensors and struggle to effectively process HD radar measurements, primarily due to memory limitations.

\textit{\textbf{Vehicle detection:}} Since our approach directly exploits point cloud information from HD radar data and is evaluated on the RADIal dataset, we adopt as baselines those architectures that are likewise trained on RADIal. As summarized in Section~\ref{relatedwork}, FFTRadNet~\cite{rebut_raw_2022}, TFFTRadNet~\cite{giroux_t-fftradnet_2023}, ADCNet~\cite{yang_adcnet_2023}, and SparseRadNet~\cite{wu_sparseradnet_2024} fall into this category. Furthermore, because our objective is to integrate camera, radar, and LiDAR modalities, we also compare against recent cross-modal fusion frameworks, including Cross-Modal Supervision (CMS)~\cite{jin_cross-modal_2023}, ROFusion~\cite{liu_rofusion_2023}, EchoFusion~\cite{liu_echoes_2023}, REFNet~\cite{chandrasekaran_resource_2024}, and REFNet++~\cite{chandrasekaran_refnet_2025} which constitute the most relevant references for benchmarking detection performance.

\textit{\textbf{Free-space segmentation:}} In the context of free-space segmentation, no prior multimodal fusion framework tailored to our setting is publicly available. We therefore reimplement the REFNet series~\cite{chandrasekaran_resource_2024, chandrasekaran_refnet_2025} and adapt it by integrating our segmentation head. In addition, we benchmark our method against other state-of-the-art approaches that operate on raw range-Doppler (RD) data from the RADIal dataset, namely SparseRadNet~\cite{wu_sparseradnet_2024}, Occugrid~\cite{jin_semantic_2024}, and TransRadar~\cite{dalbah_transradar_2024}.

% \vspace{8mm}

\subsection{Quantitative results} \label{results}
% We reimplemented all the models to compute resource efficiency as we wanted to use the same training setup and parameters from Section~\ref{traindetails} for a fair comparison.
To ensure a rigorous and fair comparison of resource efficiency, we re-implemented all models and trained them under the unified configuration and hyperparameter settings described in Section~\ref{traindetails}.

% \subsubsection{\textit{\textbf{Quantitative results}}}

\textit{\textbf{Accuracy:}} For object detection, our approach achieves superior performance in terms of Average Precision (AP) and F1-score, as reported in Table~\ref{table:quantresul}. The slight gap in Average Recall (AR) relative to ROFusion~\cite{liu_rofusion_2023} can be attributed to it's use of raw range-Doppler (RD) inputs, which improves recall at the expense of runtime efficiency (frames per second, FPS). Despite employing a substantially more complex architecture with approximately 7.8 times more trainable parameters than our model, EchoFusion~\cite{liu_echoes_2023} still underperforms our method in overall accuracy. 

Furthermore, as summarized in Table~\ref{table:quantresul}, our framework delivers markedly higher mIoU for the free-space segmentation task compared to existing methods. This improvement arises from the fact that the input data is directly ingested to the network without any intermediate physical transformations; instead, the model is trained to jointly learn the representations of image data and point cloud information in the perspective view.

\textit{\textbf{Resource efficiency:}} Our design explicitly targets a favorable trade-off between accuracy and computational overhead, yielding the second-highest FPS among all evaluated methods. Pure radar-based approaches such as FTTRadNet~\cite{rebut_raw_2022}, TFFTRadNet~\cite{giroux_t-fftradnet_2023}, and TransRadar~\cite{dalbah_transradar_2024} are still highly efficient in terms of runtime, but their reliance on a single modality limits overall accuracy. In contrast, fusion-based frameworks like ROFusion~\cite{liu_rofusion_2023} and EchoFusion~\cite{liu_echoes_2023} incur substantial GPU memory demands, indicating suboptimal resource utilization. Although REFNet~\cite{chandrasekaran_resource_2024} appears computationally efficient, it employs a separate image-processing pipeline for the camera stream whose cost is excluded from the reported complexity. REFNet++~\cite{chandrasekaran_refnet_2025}, by contrast, offers only a slight improvement. By comparison, our fusion architecture achieves a more favorable balance between performance and efficiency, attaining state-of-the-art accuracy while maintaining competitive throughput and memory usage, as summarized in Table~\ref{table:quantresul}. Furthermore, we systematically benchmark against single-modality baselines to confirm that the proposed fusion strategy remains competitive in terms of processing speed, stability, and resource consumption.
% Our design balances accuracy and efficiency, achieving the second-highest FPS while maintaining
% state-of-the-art accuracy with competitive memory use. Pure
% radar methods are fast but less accurate, and existing fusion
% models either overuse GPU memory or omit parts of their
% computational cost. As shown in Table~\ref{table:quantresul}, our fusion remains competitive in speed, stability, and resource consumption
% compared to both fusion and single-modality baselines.

% \color[HTML]{0072B2} #green
% {\color[HTML]{009E73} #blue
% {\color[HTML]{CC79A7} #orange

%%%%%%%%%%%% QUALITATIVE RESULTS %%%%%%%%%%%%%%%%%%%%%%%%%%%%%

\subsection{Qualitative results} \label{resultsquali}
\begin{figure*}[!ht]
    \centering
    \vspace{0.18cm}
    \hspace*{0.6cm}
    % Row 1
\makebox[\textwidth][c]{%
    \begin{subfigure}{0.3\textwidth}
        \includegraphics[width=4.3cm, height=2.15cm]{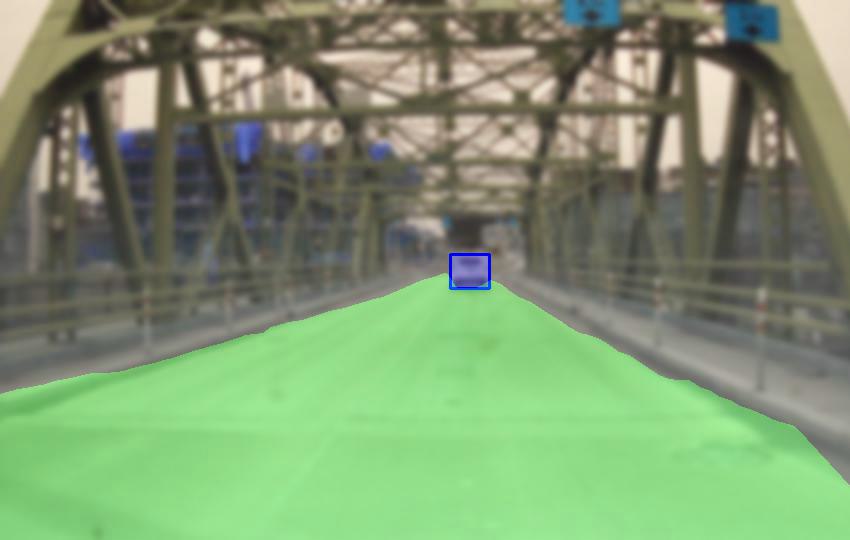}
    \end{subfigure}\hspace{-2.7em}
    \begin{subfigure}{0.3\textwidth}
        \includegraphics[width=4.3cm, height=2.15cm]{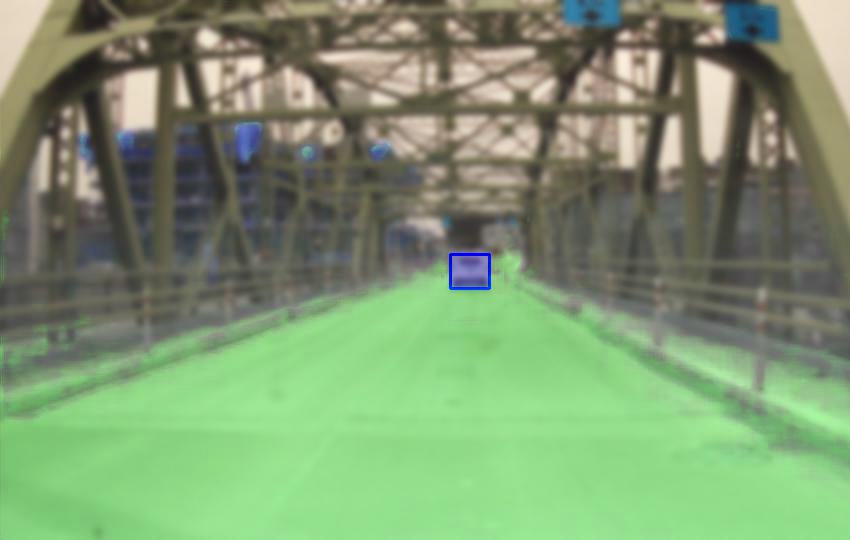}
    \end{subfigure}\hspace{-2.7em}
    \begin{subfigure}{0.3\textwidth}
        \includegraphics[width=4.3cm, height=2.15cm]{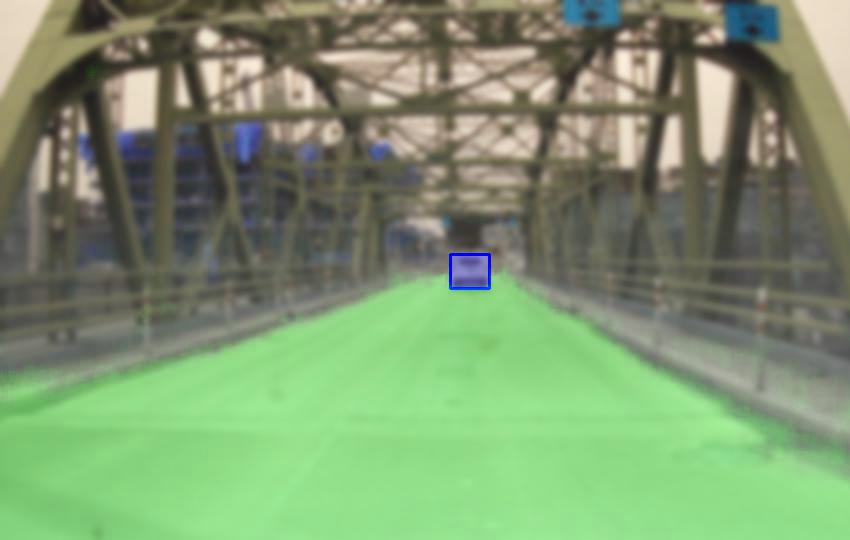}
    \end{subfigure}
}

    \vspace{0.2cm}
    \hspace*{0.6cm}
    % Row 2
\makebox[\textwidth][c]{%
    \begin{subfigure}{0.3\textwidth}
        \includegraphics[width=4.3cm, height=2.15cm]{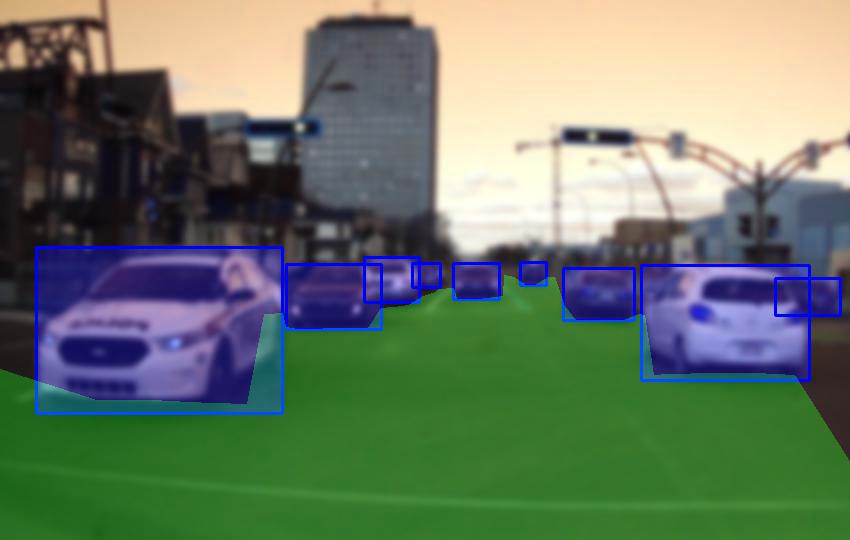}
    \end{subfigure}\hspace{-2.7em}
    \begin{subfigure}{0.3\textwidth}
        \includegraphics[width=4.3cm, height=2.15cm]{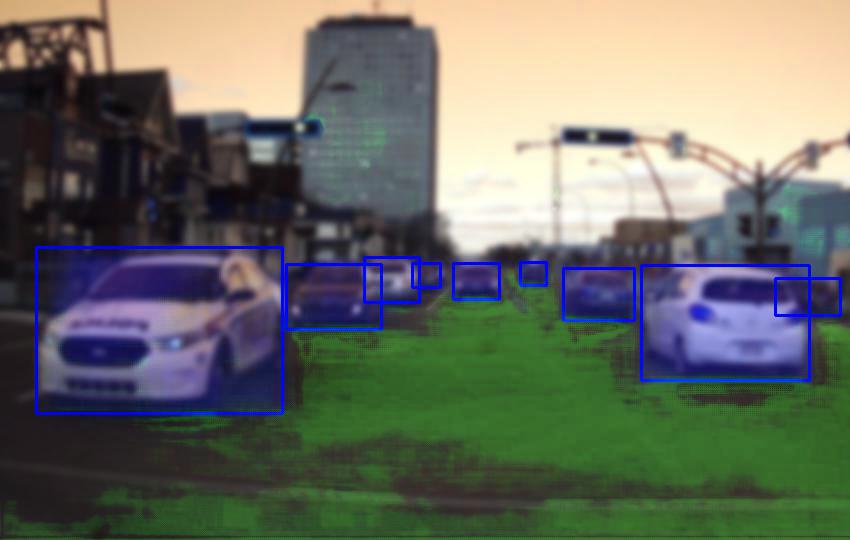}
    \end{subfigure}\hspace{-2.7em}
    \begin{subfigure}{0.3\textwidth}
        \includegraphics[width=4.3cm, height=2.15cm]{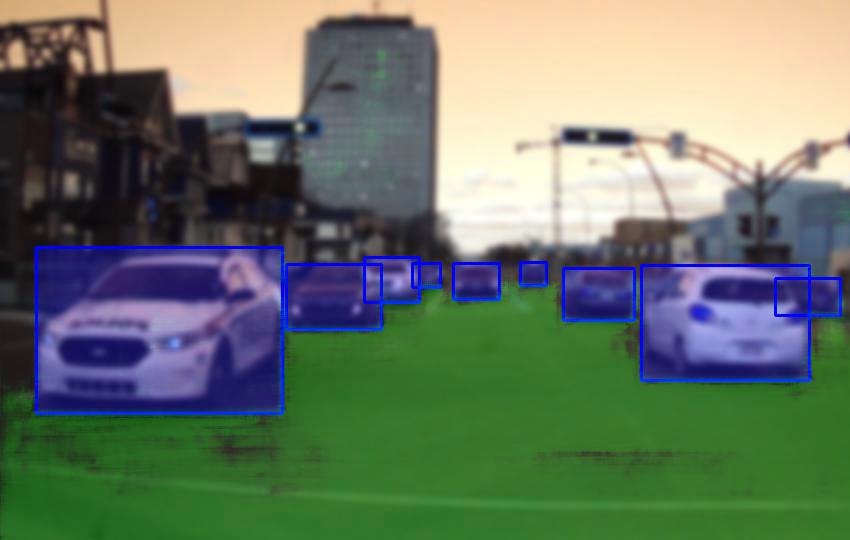}
    \end{subfigure}
}
    % Rotated text annotations using TikZ
    \begin{tikzpicture}[overlay, remember picture]
        % Annotation for the first row
        \node[rotate=0] at (-4.4, 0.15) {{ (a)}};
        \node[rotate=0] at (0.2, 0.15) {{ (b)}};
        \node[rotate=0] at (4.7, 0.15) {{ (c)}};
    \end{tikzpicture}

    \caption{Qualitative evaluation on the test set: (a) ground truth, (b) camera-only predictions, and (c) camera–radar–LiDAR fusion results. The green regions denote free-space segmentation, while the blue regions indicate detected vehicles.}
    \label{fig:qualitativeresults_fusion_new}
\end{figure*}

We qualitatively compare the fused model against the \textit{camera-only} baseline, as shown in Fig.~\ref{fig:qualitativeresults_fusion_new}, using two scenes of differing complexity. In the first scenario, the \textit{camera-only} model incorrectly classifies platform regions as free space (false positives), whereas the proposed \textit{camera-radar-LiDAR} fusion model successfully suppresses these errors. In the second, more challenging scene, the \textit{camera-only} segmentation fails to identify substantial portions of free space (false negatives), while our fusion approach provides a more complete and meaningful delineation. These observations underscore the robustness of the fusion framework and its suitability for deployment in real-world environments.

\section{CONCLUSION AND FUTURE WORK} \label{conclusion}

The current work focuses on binary classification. Future research is to extend these models to multi-class settings, which will require large-scale, high-quality, temporally synchronized multimodal datasets with precise annotations. The development of a comprehensive, diverse multimodal benchmark dataset would therefore be a critical step toward accelerating progress and enabling more rigorous evaluation of advanced fusion and multi-task learning methods.

\bibliographystyle{ieeetr}
\bibliography{thirdpub.bib}

@inproceedings{chandrasekaran_resource_2024,
	title = {A {Resource} {Efficient} {Fusion} {Network} for {Object} {Detection} in {Bird}'s-{Eye} {View} using {Camera} and {Raw} {Radar} {Data}},
	issn = {2153-0017},
	url = {https://ieeexplore.ieee.org/document/10919729},
	doi = {10.1109/ITSC58415.2024.10919729},
	urldate = {2026-03-08},
	booktitle = {2024 {IEEE} 27th {International} {Conference} on {Intelligent} {Transportation} {Systems} ({ITSC})},
	author = {Chandrasekaran, Kavin and Grigorescu, Sorin and Dubbelman, Gijs and Jancura, Pavol},
	month = sep,
	year = {2024},
	note = {ISSN: 2153-0017},
	pages = {41--48},
}

@misc{caesar_nuscenes_2020,
	title = {{nuScenes}: {A} multimodal dataset for autonomous driving},
	shorttitle = {{nuScenes}},
	url = {http://arxiv.org/abs/1903.11027},
	doi = {10.48550/arXiv.1903.11027},
	urldate = {2023-12-19},
	publisher = {arXiv},
	author = {Caesar, Holger and Bankiti, Varun and Lang, Alex H. and Vora, Sourabh and Liong, Venice Erin and Xu, Qiang and Krishnan, Anush and Pan, Yu and Baldan, Giancarlo and Beijbom, Oscar},
	month = may,
	year = {2020},
	note = {arXiv:1903.11027 [cs, stat]},
}

@misc{chen_futr3d_2023,
	title = {{FUTR3D}: {A} {Unified} {Sensor} {Fusion} {Framework} for {3D} {Detection}},
	shorttitle = {{FUTR3D}},
	url = {http://arxiv.org/abs/2203.10642},
	doi = {10.48550/arXiv.2203.10642},
	urldate = {2024-02-01},
	publisher = {arXiv},
	author = {Chen, Xuanyao and Zhang, Tianyuan and Wang, Yue and Wang, Yilun and Zhao, Hang},
	month = apr,
	year = {2023},
	note = {arXiv:2203.10642 [cs]},
}

@misc{chandrasekaran_refnet_2025,
	title = {{REFNet}++: {Multi}-{Task} {Efficient} {Fusion} of {Camera} and {Radar} {Sensor} {Data} in {Bird}’s-{Eye} {Polar} {View}},
	url = {https://github.com/tue-mps/refnetplusplus},
	urldate = {2026-02-20},
	publisher = {2025 IEEE 28th International Conference on Intelligent Transportation Systems (ITSC), to appear in IEEE Xplore.},
	author = {Chandrasekaran, Kavin and Grigorescu, Sorin},
	month = jul,
	year = {2025},
	note = {2025 IEEE 28th International Conference on Intelligent Transportation Systems (ITSC), to appear in IEEE Xplore.},
}

@misc{cai_bevfusion4d_2023,
	title = {{BEVFusion4D}: {Learning} {LiDAR}-{Camera} {Fusion} {Under} {Bird}'s-{Eye}-{View} via {Cross}-{Modality} {Guidance} and {Temporal} {Aggregation}},
	shorttitle = {{BEVFusion4D}},
	url = {http://arxiv.org/abs/2303.17099},
	doi = {10.48550/arXiv.2303.17099},
	urldate = {2025-12-22},
	publisher = {arXiv},
	author = {Cai, Hongxiang and Zhang, Zeyuan and Zhou, Zhenyu and Li, Ziyin and Ding, Wenbo and Zhao, Jiuhua},
	month = mar,
	year = {2023},
	note = {arXiv:2303.17099 [cs]},
}

@misc{xia_insfusion_2025,
	title = {{InsFusion}: {Rethink} {Instance}-level {LiDAR}-{Camera} {Fusion} for {3D} {Object} {Detection}},
	shorttitle = {{InsFusion}},
	url = {http://arxiv.org/abs/2509.08374},
	doi = {10.48550/arXiv.2509.08374},
	urldate = {2025-12-22},
	publisher = {arXiv},
	author = {Xia, Zhongyu and Yang, Hansong and Wang, Yongtao},
	month = nov,
	year = {2025},
	note = {arXiv:2509.08374 [cs]},
}

@misc{noauthor_opencv_nodate,
	title = {{OpenCV}: {Camera} {Calibration} and {3D} {Reconstruction}},
	url = {https://docs.opencv.org/3.4/d9/d0c/group__calib3d.html#ga1019495a2c8d1743ed5cc23fa0daff8c},
	urldate = {2025-12-15},
}

@misc{noauthor_rotation_nodate,
	title = {Rotation — {SciPy} v1.16.2 {Manual}},
	url = {https://docs.scipy.org/doc//scipy-1.16.2/reference/generated/scipy.spatial.transform.Rotation.html},
	urldate = {2025-12-15},
}

@article{long_radiant_2023,
	title = {{RADIANT}: {Radar}-{Image} {Association} {Network} for {3D} {Object} {Detection}},
	volume = {37},
	copyright = {Copyright (c) 2023 Association for the Advancement of Artificial Intelligence},
	issn = {2374-3468},
	shorttitle = {{RADIANT}},
	url = {https://ojs.aaai.org/index.php/AAAI/article/view/25270},
	doi = {10.1609/aaai.v37i2.25270},
	language = {en},
	number = {2},
	urldate = {2025-04-09},
	journal = {Proceedings of the AAAI Conference on Artificial Intelligence},
	author = {Long, Yunfei and Kumar, Abhinav and Morris, Daniel and Liu, Xiaoming and Castro, Marcos and Chakravarty, Punarjay},
	month = jun,
	year = {2023},
	note = {Number: 2},
	pages = {1808--1816},
}

@misc{lei_hvdetfusion_2023,
	title = {{HVDetFusion}: {A} {Simple} and {Robust} {Camera}-{Radar} {Fusion} {Framework}},
	shorttitle = {{HVDetFusion}},
	url = {http://arxiv.org/abs/2307.11323},
	doi = {10.48550/arXiv.2307.11323},
	urldate = {2025-04-09},
	publisher = {arXiv},
	author = {Lei, Kai and Chen, Zhan and Jia, Shuman and Zhang, Xiaoteng},
	month = jul,
	year = {2023},
	note = {arXiv:2307.11323 [cs]},
}

@misc{kim_rcm-fusion_2024,
	title = {{RCM}-{Fusion}: {Radar}-{Camera} {Multi}-{Level} {Fusion} for {3D} {Object} {Detection}},
	shorttitle = {{RCM}-{Fusion}},
	url = {http://arxiv.org/abs/2307.10249},
	doi = {10.48550/arXiv.2307.10249},
	urldate = {2025-04-09},
	publisher = {arXiv},
	author = {Kim, Jisong and Seong, Minjae and Bang, Geonho and Kum, Dongsuk and Choi, Jun Won},
	month = may,
	year = {2024},
	note = {arXiv:2307.10249 [cs]},
}

@misc{shi_radar_2024,
	title = {Radar and {Camera} {Fusion} for {Object} {Detection} and {Tracking}: {A} {Comprehensive} {Survey}},
	shorttitle = {Radar and {Camera} {Fusion} for {Object} {Detection} and {Tracking}},
	url = {http://arxiv.org/abs/2410.19872},
	doi = {10.48550/arXiv.2410.19872},
	urldate = {2025-04-07},
	publisher = {arXiv},
	author = {Shi, Kun and He, Shibo and Shi, Zhenyu and Chen, Anjun and Xiong, Zehui and Chen, Jiming and Luo, Jun},
	month = oct,
	year = {2024},
	note = {arXiv:2410.19872 [cs]},
}

@article{roldan_deep_2024,
	title = {A {Deep} {Automotive} {Radar} {Detector} {Using} the {RaDelft} {Dataset}},
	volume = {2},
	issn = {2832-7357},
	url = {https://ieeexplore.ieee.org/document/10731871},
	doi = {10.1109/TRS.2024.3485578},
	urldate = {2025-04-06},
	journal = {IEEE Transactions on Radar Systems},
	author = {Roldan, Ignacio and Palffy, Andras and Kooij, Julian F. P. and Gavrila, Dariu M. and Fioranelli, Francesco and Yarovoy, Alexander},
	year = {2024},
	pages = {1062--1075},
}

@inproceedings{rebut_raw_2022,
	title = {Raw {High}-{Definition} {Radar} for {Multi}-{Task} {Learning}},
	url = {https://openaccess.thecvf.com/content/CVPR2022/html/Rebut_Raw_High-Definition_Radar_for_Multi-Task_Learning_CVPR_2022_paper.html},
	language = {en},
	urldate = {2025-04-06},
	booktitle = {{IEEE}/{CVF} {Conference} on {Computer} {Vision} and {Pattern} {Recognition}},
	publisher = {IEEE},
	author = {Rebut, Julien and Ouaknine, Arthur and Malik, Waqas and Pérez, Patrick},
	year = {2022},
	pages = {17021--17030},
}

@misc{wu_sparseradnet_2024,
	title = {{SparseRadNet}: {Sparse} {Perception} {Neural} {Network} on {Subsampled} {Radar} {Data}},
	shorttitle = {{SparseRadNet}},
	url = {http://arxiv.org/abs/2406.10600},
	doi = {10.48550/arXiv.2406.10600},
	urldate = {2025-01-17},
	publisher = {arXiv},
	author = {Wu, Jialong and Meuter, Mirko and Schoeler, Markus and Rottmann, Matthias},
	month = jul,
	year = {2024},
	note = {arXiv:2406.10600 [cs]},
}

@article{jin_semantic_2024,
	title = {Semantic {Segmentation}-{Based} {Occupancy} {Grid} {Map} {Learning} {With} {Automotive} {Radar} {Raw} {Data}},
	volume = {9},
	issn = {2379-8904},
	url = {https://ieeexplore.ieee.org/document/10273590},
	doi = {10.1109/TIV.2023.3322353},
	number = {1},
	urldate = {2025-01-16},
	journal = {IEEE Transactions on Intelligent Vehicles},
	author = {Jin, Yi and Hoffmann, Marcel and Deligiannis, Anastasios and Fuentes-Michel, Juan-Carlos and Vossiek, Martin},
	month = jan,
	year = {2024},
	note = {Conference Name: IEEE Transactions on Intelligent Vehicles},
	pages = {216--230},
}

@inproceedings{dalbah_transradar_2024,
	address = {Waikoloa, HI, USA},
	title = {{TransRadar}: {Adaptive}-{Directional} {Transformer} for {Real}-{Time} {Multi}-{View} {Radar} {Semantic} {Segmentation}},
	copyright = {https://doi.org/10.15223/policy-029},
	isbn = {979-8-3503-1892-0},
	shorttitle = {{TransRadar}},
	url = {https://ieeexplore.ieee.org/document/10483848/},
	doi = {10.1109/WACV57701.2024.00042},
	language = {en},
	urldate = {2025-01-16},
	booktitle = {2024 {IEEE}/{CVF} {Winter} {Conference} on {Applications} of {Computer} {Vision} ({WACV})},
	publisher = {IEEE},
	author = {Dalbah, Yahia and Lahoud, Jean and Cholakkal, Hisham},
	month = jan,
	year = {2024},
	pages = {352--361},
}

@article{lu_monocular_2019,
	title = {Monocular {Semantic} {Occupancy} {Grid} {Mapping} with {Convolutional} {Variational} {Encoder}-{Decoder} {Networks}},
	volume = {4},
	issn = {2377-3766, 2377-3774},
	url = {http://arxiv.org/abs/1804.02176},
	doi = {10.1109/LRA.2019.2891028},
	number = {2},
	urldate = {2025-01-04},
	journal = {IEEE Robotics and Automation Letters},
	author = {Lu, Chenyang and Molengraft, Marinus Jacobus Gerardus van de and Dubbelman, Gijs},
	month = apr,
	year = {2019},
	note = {arXiv:1804.02176 [cs]},
	pages = {445--452},
}

@misc{yang_adcnet_2023,
	title = {{ADCNet}: {Learning} from {Raw} {Radar} {Data} via {Distillation}},
	shorttitle = {{ADCNet}},
	url = {http://arxiv.org/abs/2303.11420},
	doi = {10.48550/arXiv.2303.11420},
	urldate = {2024-03-29},
	publisher = {arXiv},
	author = {Yang, Bo and Khatri, Ishan and Happold, Michael and Chen, Chulong},
	month = dec,
	year = {2023},
	note = {arXiv:2303.11420 [cs, eess]},
}

@inproceedings{man_bev-guided_2023,
	title = {{BEV}-{Guided} {Multi}-{Modality} {Fusion} for {Driving} {Perception}},
	url = {https://openaccess.thecvf.com/content/CVPR2023/html/Man_BEV-Guided_Multi-Modality_Fusion_for_Driving_Perception_CVPR_2023_paper.html},
	language = {en},
	urldate = {2023-12-19},
	booktitle = {{IEEE}/{CVF} {Conference} on {Computer} {Vision} and {Pattern} {Recognition} ({CVPR})},
	author = {Man, Yunze and Gui, Liang-Yan and Wang, Yu-Xiong},
	year = {2023},
	pages = {21960--21969},
}

@article{xiong_lxl_2023,
	title = {{LXL}: {LiDAR} {Excluded} {Lean} {3D} {Object} {Detection} with {4D} {Imaging} {Radar} and {Camera} {Fusion}},
	issn = {2379-8904, 2379-8858},
	shorttitle = {{LXL}},
	url = {https://ieeexplore.ieee.org/document/10268601/},
	doi = {10.1109/TIV.2023.3321240},
	urldate = {2023-12-19},
	journal = {IEEE Transactions on Intelligent Vehicles},
	author = {Xiong, Weiyi and Liu, Jianan and Huang, Tao and Han, Qing-Long and Xia, Yuxuan and Zhu, Bing},
	year = {2023},
	pages = {1--14},
}

@article{rizzoli_multimodal_2022,
	title = {Multimodal {Semantic} {Segmentation} in {Autonomous} {Driving}: {A} {Review} of {Current} {Approaches} and {Future} {Perspectives}},
	volume = {10},
	issn = {2227-7080},
	shorttitle = {Multimodal {Semantic} {Segmentation} in {Autonomous} {Driving}},
	url = {https://www.mdpi.com/2227-7080/10/4/90},
	doi = {10.3390/technologies10040090},
	language = {en},
	number = {4},
	urldate = {2023-12-16},
	journal = {Technologies},
	author = {Rizzoli, Giulia and Barbato, Francesco and Zanuttigh, Pietro},
	month = jul,
	year = {2022},
	pages = {90},
}

@article{zheng_rcfusion_2023,
	title = {{RCFusion}: {Fusing} 4-{D} {Radar} and {Camera} {With} {Bird}’s-{Eye} {View} {Features} for 3-{D} {Object} {Detection}},
	volume = {72},
	issn = {0018-9456, 1557-9662},
	shorttitle = {{RCFusion}},
	url = {https://ieeexplore.ieee.org/document/10138035/},
	doi = {10.1109/TIM.2023.3280525},
	urldate = {2023-12-16},
	journal = {IEEE Transactions on Instrumentation and Measurement},
	author = {Zheng, Lianqing and Li, Sen and Tan, Bin and Yang, Long and Chen, Sihan and Huang, Libo and Bai, Jie and Zhu, Xichan and Ma, Zhixiong},
	year = {2023},
	pages = {1--14},
}

@article{liu_echoes_2023,
	title = {Echoes {Beyond} {Points}: {Unleashing} the {Power} of {Raw} {Radar} {Data} in {Multi}-modality {Fusion}},
	volume = {36},
	shorttitle = {Echoes {Beyond} {Points}},
	url = {https://proceedings.neurips.cc/paper_files/paper/2023/hash/a8f7f12b29d9b8c227785f6b529f63b7-Abstract-Conference.html},
	language = {en},
	urldate = {2024-03-22},
	journal = {Advances in Neural Information Processing Systems},
	author = {Liu, Yang and Wang, Feng and Wang, Naiyan and Zhang, Zhao-Xiang},
	month = dec,
	year = {2023},
	pages = {53964--53982},
}

@article{jin_cross-modal_2023,
	title = {Cross-{Modal} {Supervision}-{Based} {Multitask} {Learning} {With} {Automotive} {Radar} {Raw} {Data}},
	volume = {8},
	issn = {2379-8904},
	url = {https://ieeexplore.ieee.org/document/10008067},
	doi = {10.1109/TIV.2023.3234583},
	number = {4},
	urldate = {2024-01-30},
	journal = {IEEE Transactions on Intelligent Vehicles},
	author = {Jin, Yi and Deligiannis, Anastasios and Fuentes-Michel, Juan-Carlos and Vossiek, Martin},
	month = apr,
	year = {2023},
	note = {Conference Name: IEEE Transactions on Intelligent Vehicles},
	pages = {3012--3025},
}

@misc{lin_focal_2018,
	title = {Focal {Loss} for {Dense} {Object} {Detection}},
	url = {http://arxiv.org/abs/1708.02002},
	doi = {10.48550/arXiv.1708.02002},
	urldate = {2024-03-20},
	publisher = {arXiv},
	author = {Lin, Tsung-Yi and Goyal, Priya and Girshick, Ross and He, Kaiming and Dollár, Piotr},
	month = feb,
	year = {2018},
	note = {arXiv:1708.02002 [cs]},
}

@misc{janai_computer_2021,
	title = {Computer {Vision} for {Autonomous} {Vehicles}: {Problems}, {Datasets} and {State} of the {Art}},
	shorttitle = {Computer {Vision} for {Autonomous} {Vehicles}},
	url = {http://arxiv.org/abs/1704.05519},
	doi = {10.48550/arXiv.1704.05519},
	urldate = {2024-02-01},
	publisher = {arXiv},
	author = {Janai, Joel and Güney, Fatma and Behl, Aseem and Geiger, Andreas},
	month = mar,
	year = {2021},
	note = {arXiv:1704.05519 [cs]},
}

@misc{giroux_t-fftradnet_2023,
	title = {T-{FFTRadNet}: {Object} {Detection} with {Swin} {Vision} {Transformers} from {Raw} {ADC} {Radar} {Signals}},
	shorttitle = {T-{FFTRadNet}},
	url = {http://arxiv.org/abs/2303.16940},
	doi = {10.48550/arXiv.2303.16940},
	urldate = {2024-01-30},
	publisher = {arXiv},
	author = {Giroux, James and Bouchard, Martin and Laganiere, Robert},
	month = mar,
	year = {2023},
	note = {arXiv:2303.16940 [cs]},
}

@misc{hendy_fishing_2020,
	title = {{FISHING} {Net}: {Future} {Inference} of {Semantic} {Heatmaps} {In} {Grids}},
	shorttitle = {{FISHING} {Net}},
	url = {http://arxiv.org/abs/2006.09917},
	urldate = {2023-12-18},
	publisher = {arXiv},
	author = {Hendy, Noureldin and Sloan, Cooper and Tian, Feng and Duan, Pengfei and Charchut, Nick and Xie, Yuesong and Wang, Chuang and Philbin, James},
	month = jun,
	year = {2020},
	note = {arXiv:2006.09917 [cs]},
}

@misc{drews_deepfusion_2022,
	title = {{DeepFusion}: {A} {Robust} and {Modular} {3D} {Object} {Detector} for {Lidars}, {Cameras} and {Radars}},
	shorttitle = {{DeepFusion}},
	url = {http://arxiv.org/abs/2209.12729},
	doi = {10.48550/arXiv.2209.12729},
	urldate = {2023-12-16},
	publisher = {arXiv},
	author = {Drews, Florian and Feng, Di and Faion, Florian and Rosenbaum, Lars and Ulrich, Michael and Gläser, Claudius},
	month = sep,
	year = {2022},
	note = {arXiv:2209.12729 [cs]},
}

@misc{bansal_radsegnet_2022,
	title = {{RadSegNet}: {A} {Reliable} {Approach} to {Radar} {Camera} {Fusion}},
	shorttitle = {{RadSegNet}},
	url = {http://arxiv.org/abs/2208.03849},
	urldate = {2023-12-16},
	publisher = {arXiv},
	author = {Bansal, Kshitiz and Rungta, Keshav and Bharadia, Dinesh},
	month = aug,
	year = {2022},
	note = {arXiv:2208.03849 [cs]},
}

@misc{zhao_lif-seg_2021,
	title = {{LIF}-{Seg}: {LiDAR} and {Camera} {Image} {Fusion} for {3D} {LiDAR} {Semantic} {Segmentation}},
	shorttitle = {{LIF}-{Seg}},
	url = {http://arxiv.org/abs/2108.07511},
	doi = {10.48550/arXiv.2108.07511},
	urldate = {2023-12-16},
	publisher = {arXiv},
	author = {Zhao, Lin and Zhou, Hui and Zhu, Xinge and Song, Xiao and Li, Hongsheng and Tao, Wenbing},
	month = aug,
	year = {2021},
	note = {arXiv:2108.07511 [cs]},
}

@misc{paek_k-radar_2023,
	title = {K-{Radar}: {4D} {Radar} {Object} {Detection} for {Autonomous} {Driving} in {Various} {Weather} {Conditions}},
	shorttitle = {K-{Radar}},
	url = {http://arxiv.org/abs/2206.08171},
	doi = {10.48550/arXiv.2206.08171},
	urldate = {2023-12-14},
	publisher = {arXiv},
	author = {Paek, Dong-Hee and Kong, Seung-Hyun and Wijaya, Kevin Tirta},
	month = nov,
	year = {2023},
	note = {arXiv:2206.08171 [cs]},
}

@misc{kingma_adam_2017,
	title = {Adam: {A} {Method} for {Stochastic} {Optimization}},
	shorttitle = {Adam},
	url = {http://arxiv.org/abs/1412.6980},
	doi = {10.48550/arXiv.1412.6980},
	urldate = {2023-12-14},
	publisher = {arXiv},
	author = {Kingma, Diederik P. and Ba, Jimmy},
	month = jan,
	year = {2017},
	note = {arXiv:1412.6980 [cs]},
}

@misc{liu_rofusion_2023,
	title = {{ROFusion}: {Efficient} {Object} {Detection} using {Hybrid} {Point}-wise {Radar}-{Optical} {Fusion}},
	shorttitle = {{ROFusion}},
	url = {http://arxiv.org/abs/2307.08233},
	doi = {10.48550/arXiv.2307.08233},
	urldate = {2023-12-14},
	publisher = {arXiv},
	author = {Liu, Liu and Zhi, Shuaifeng and Du, Zhenhua and Liu, Li and Zhang, Xinyu and Huo, Kai and Jiang, Weidong},
	month = jul,
	year = {2023},
	note = {arXiv:2307.08233 [cs]},
}

\end{document}